\documentclass[]{spie}  

\usepackage{amsmath,amsfonts,amssymb}
\usepackage{graphicx}
\usepackage{times}
\usepackage{tabularx}
\usepackage[normalem]{ulem}
\usepackage{adjustbox}
\usepackage[colorlinks=true, allcolors=blue]{hyperref}
\newcommand{\xxx}[1]{ \textcolor{red}{#1} }

\newcolumntype{P}[1]{>{\centering\arraybackslash}p{#1}}
\newcolumntype{M}[1]{>{\centering\arraybackslash}m{#1}}

\title{False Negative/Positive Control for SAM on Noisy Medical Images}
\author[a]{Xing Yao}
\author[a]{Han Liu}
\author[b]{Dewei Hu}
\author[a]{Daiwei Lu}
\author[b]{Ange Lou}
\author[b]{Hao Li}
\author[a]{Ruining Deng}
\author[d]{Gabriel Arenas}
\author[d]{Baris Oguz}
\author[d]{Nadav Schwartz}
\author[c]{Brett C Byram}
\author[a]{Ipek Oguz}
\affil[a]{Dept.\ of Computer Science, Vanderbilt University, Nashville, TN, USA} 
\affil[b]{Dept.\ of Electrical and Computer Engineering, Vanderbilt University, Nashville, TN, USA}
\affil[c]{Dept.\ of Biomedical Engineering, Vanderbilt University, Nashville, TN, USA}
\affil[d]{Dept.\ of Obstetrics and Gynecology,
University of Pennsylvania, Philadelphia, PA, USA}

\begin{document} 
\maketitle

\begin{abstract}
The Segment Anything Model (SAM) is a recently developed all-range foundation model for image segmentation. It can use sparse manual prompts such as bounding boxes to generate pixel-level segmentation in natural images but struggles in medical images such as low-contrast, noisy ultrasound images. We propose a refined test-phase prompt augmentation technique designed to improve SAM's performance in medical image segmentation. The method couples multi-box prompt augmentation and an aleatoric uncertainty-based false-negative (FN) and false-positive (FP) correction (FNPC) strategy. We evaluate the method on two ultrasound datasets and show improvement in SAM's performance and robustness to inaccurate prompts, without the necessity for further training or tuning. Moreover, we present the Single-Slice-to-Volume (SS2V) method, enabling 3D pixel-level segmentation using only the bounding box annotation from a single 2D slice. Our results allow efficient use of SAM in even noisy, low-contrast medical images. The source code will be released soon at: {\url{https://github.com/xyimaging/FNPC}}


\end{abstract}


\section{Introduction}
\label{sec:intro}  
The performance of deep convolutional neural networks (DNNs) relies on large datasets with pixel-level annotations, which can be both time-consuming and labor-intensive to obtain. To address this, researchers have explored coarse-to-fine segmentation, learning pixel-level masks from easier-to-obtain annotations like bounding boxes (BBs) \cite{xu2021box}. 

Recently, the Segment Anything Model (SAM)\cite{kirillov2023segment} proposed by Meta AI has gained significant attention as an all-range segmentation foundation model, capable of generating fine-grade segmentation masks using sparse prompts like BBs, fore/background points, texts, and dense prompts such as masks. While SAM excels in natural image segmentation\cite{yang2023track, yu2023inpaint,li2023semantic}, its performance in various medical image segmentation tasks can be unsatisfactory \cite{deng2023segment, he2023accuracy, roy2023sam, mattjie2023exploring}, leading to increased interest in enhancing its capabilities in medical imaging scenarios\cite{wu2023medical,zhang2023input,cui2023all,shen2023temporally, zhang2023customized,deng2023sam}.

Efforts to improve SAM's performance on medical images  fall into three categories: 1) Fine-tuning a modified SAM model with extensive medical image datasets\cite{wu2023medical,zhang2023customized}. This requires additional time and effort for annotation and training, and the results depend on data availability and task complexity. 2) Directly using SAM's predicted results as fake ground truth to train a new network\cite{cui2023all}. This approach relies on SAM's initial performance on the specific task. 3) Leveraging prompt optimization or augmentation strategies to achieve zero-shot segmentation\cite{shen2023temporally,zhang2023input, deng2023sam}.
These strategies can efficiently enhance zero-shot segmentation performance of SAM, with potential to be leveraged across unseen datasets and downstream tasks.

Inspired by \cite{deng2023sam, shin2023sdc}, we propose a test-phase prompt augmentation method to amplify the coarse-to-fine zero-shot segmentation performance of SAM on low-contrast and noisy ultrasound images, without supplementary training or fine-tuning. We assess the  performance on kidney and placenta ultrasound images. Our results indicate that our proposed method surpasses the performance of standalone SAM and, notably, exhibits exceptional robustness to variations in the prompt. 

\begin{figure}[t]
\centering
\includegraphics[width=0.91\textwidth]{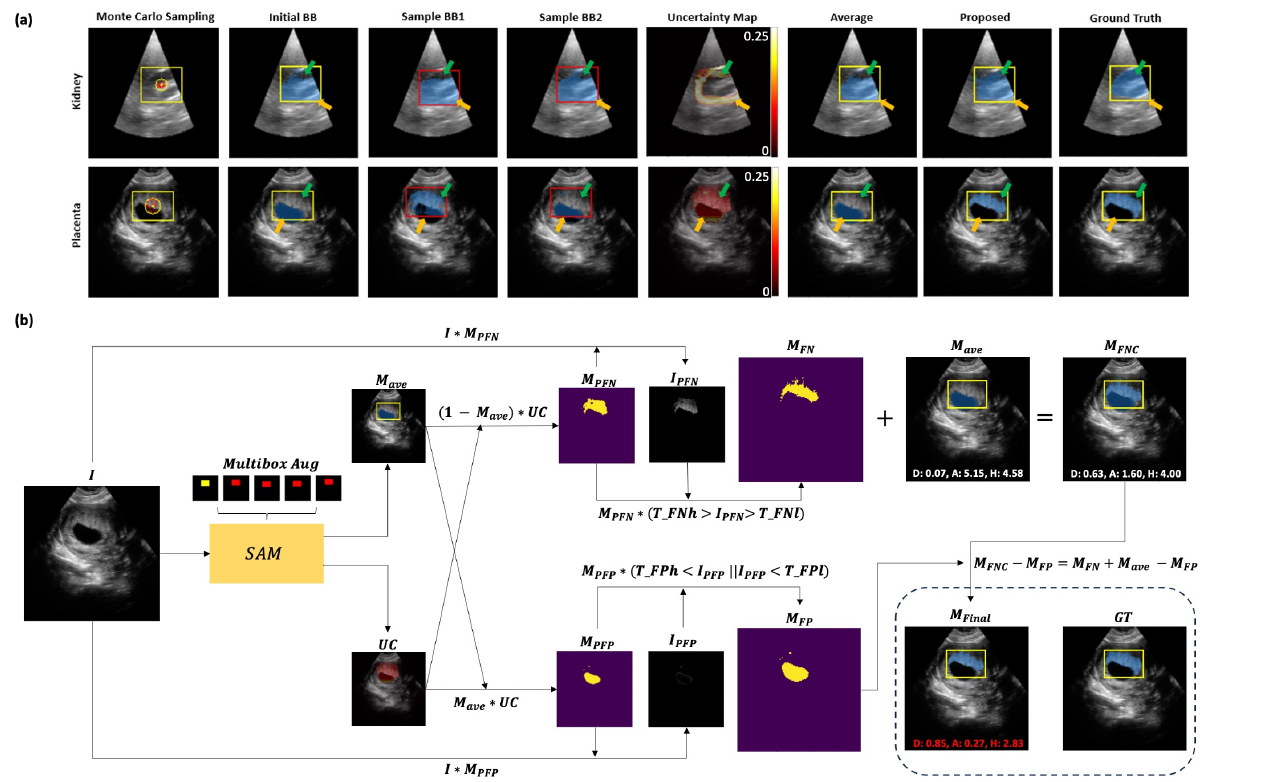}
\caption{\textbf{(a)} Monte Carlo BB sampling strategy (Sec.\ \ref{sec:mc}) on two segmentation tasks. The initial BB (yellow) has a center (yellow dot) and a sampling range (yellow circle). Sampled boxes (red) with their centers (red dots) are used to generate an average prediction. Green arrows point to FN areas in results from only the initial BB or simple averaging method, and are corrected by the proposed FNPC strategy. Orange arrows indicate areas where FPs are similarly corrected by the proposed FNPC. \textbf{(b)}  FNPC pipeline (Sec.\ \ref{sec:FNPC}).} 
\label{pipline}
\end{figure}

We have 4 main contributions: \textbf{1)} SAM's performance is sensitive to the BB position and size\cite{cheng2023sam, deng2023segment, deng2023sam}, and prediction based on a single prompt may contain FP and FN regions, as shown in Fig.\ \ref{pipline}(a). We suggest a Monte Carlo BB sampling approach to provide additional foreground/background prompts to SAM. The predictions from different field-of-view (FOV) allow us to estimate the aleatoric uncertainty map\cite{kendall2017uncertainties}. \textbf{2)} In previous studies, the uncertainty map is only used to represent the reliability and robustness of segmentation. Here, we further leverage the aleatoric uncertainty map for FN and FP correction (FNPC) in the averaged predictions. \textbf{3)} We examine the influence of the BB prompts in our study in fine (tight) BB, medium BB, and coarse BB scenarios. \textbf{4)} We introduce a Single-Slice-to-Volume (SS2V) approach for extension to 3D. This allows for pixel-level segmentation of an entire 3D volume based solely on the BB annotation of a single 2D slice.

\section{Methods}
\label{sec:method}

\subsection{Monte Carlo bounding box sampling strategy}
\label{sec:mc}
Combining SAM with bounding boxes as prompts has shown good segmentation results in medical images, although the performance is sensitive to the BB position and size\cite{cheng2023sam, deng2023segment, deng2023sam}. SAM-U\cite{deng2023sam} addresses this by implementing a multi-BB prompt augmentation approach. While SAM-U only uses a simple random sampling strategy, we propose a Monte Carlo sampling method with constrained positions, as shown in Fig.\ \ref{pipline}(a). Our sampling method involves randomly selecting $N$ points within a radius $R = \frac{1}{M} \min(\text{BB edges})$ around the center of the initial BB to  generate new BBs of the same size. $M$ is the radius ratio. This sampling strategy serves two purposes: first, the new sampled BBs (Fig.\ \ref{pipline}(a)) provide additional positive and negative prompts to SAM.
Second, our sampling approach emulates the imprecision of typical manual BB placement. 

The predictions $M_i$ from augmented BBs are averaged as $\MakeUppercase{M}_{ave} = \{\frac{1}{N + 1}\sum_{i=1}^{N + 1} M_{i}\} > T_{ave}$. While SAM-U uses a threshold of $T_{ave}=0$, effectively taking the union of the predictions, we use $T_{ave}=0.5$ for a majority vote approach. 

\subsection{Uncertainty Estimation}
\label{uc}
We propose a straightforward approach to compute aleatoric uncertainty\cite{kendall2017uncertainties}  $\MakeUppercase{UC_{raw}}$ from the  $\MakeUppercase{N + 1}$ predictions of augmented BBs and the initial BB. 
By analyzing the frequency $f$ of foreground pixels in the set of predicted masks $\{M_i\}$, the uncertainty associated with each pixel in position $(j, k)$ is determined by $
f(j, k) = \frac{1}{N + 1}\sum_{i=1}^{N + 1} M_{i}(j, k)$.
This frequency calculation allows us to compute the aleatoric uncertainty $\MakeUppercase{UC}$ (we drop the pixel coordinates from the notation for brevity): 
\begin{equation}
UC_{raw} = f \cdot (1 - f)
\label{eq2}
\end{equation}
For a more entropy-like uncertainty, we propose the following calculation, with $\epsilon=10^{-7}$:
\begin{equation}
UC_{raw} = -0.5 \cdot [f \cdot \log(f + \epsilon) + (1 - f) \cdot \log(1 - f + \epsilon)]
\label{eq3}
\end{equation}
In our experiments, Equations \ref{eq2} and \ref{eq3} share the same performance. Finally, we threshold  to extract the high-uncertainty areas: $\MakeUppercase{UC} = UC_{raw} > [min(UC_{raw}) + T_{UC} \cdot (max(UC_{raw}) - min(UC_{raw}))] $, where $T_{UC}$ is the threshold ratio. 

\subsection{False Negative and False Positive Correction (FNPC)}
\label{sec:FNPC}

Fig.\ \ref{pipline}(b) illustrates the pipeline of our proposed FNPC strategy. Given an input image $I$, we determine the average prediction $M_{ave}$ (Sec.~\ref{sec:mc}) and the uncertainty map $UC$ which highlights potential FNs and FPs (Sec.~\ref{uc}). 

\noindent
\textbf{\underline{False Negative Correction:}} Our goal is to identify FNs that are outside $M_{ave}$ but within $UC$. Initially, the potential FN mask $M_{PFN}$ is computed as $(1 - M_{ave}) \cdot UC$, and its corresponding regions in $I$ are extracted as $I_{PFN} = I \cdot M_{PFN}$. To encourage intensity homogeneity, only pixels in $I_{PFN}$ with intensities within a range $[T_{FNl}, T_{FNh}]$ are kept in the final FN mask $M_{FN}$: $M_{FN} = M_{PFN} \cdot (T_{FNh} > I_{PFN} > T_{FNl})$. 

\noindent
\textbf{\underline{False Positive Correction:}} We aim to identify FPs present in both $M_{ave}$ and $UC$. The potential FP mask $M_{PFP}$ is derived from $M_{ave} \cdot UC$, with  associated intensity values $I_{PFP} = I \cdot M_{PFP}$. The final FP mask $M_{FP}$ is again determined with an intensity range $[T_{FPl}, T_{FPh}]$:   $M_{FP} = M_{PFP} \cdot (T_{FPh} < I_{PFP} || I_{PFP} < T_{FPl})$. 

The final mask $M_{Final}$ which corrects for FNs and FPs is given by: $M_{FNPC} = M_{ave} + M_{FN} - M_{FP}$. As shown in Fig.\ \ref{pipline}(a), the FNPC has good FN and FP correction performance compared to both the raw SAM and simple averaging.

\subsection{Single Slice to Volume method (SS2V)}
\label{ss2v_sec}

\begin{figure}[h]
\centering
\includegraphics[width=0.9\textwidth]{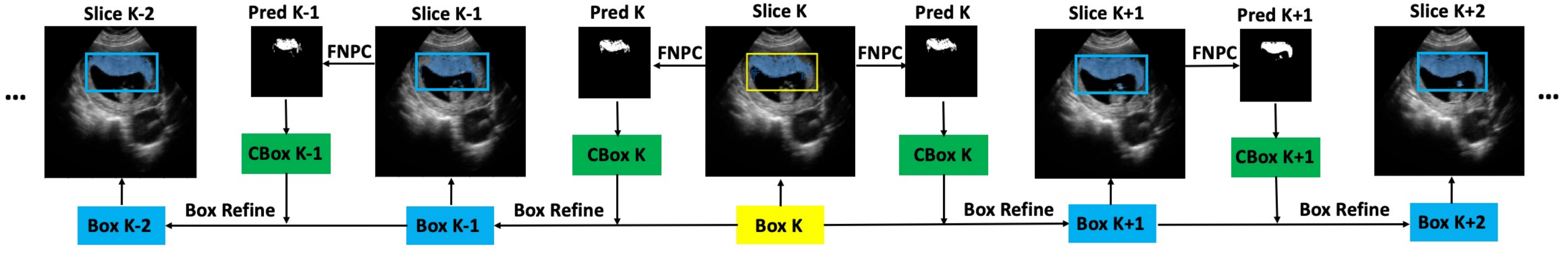}
\caption{Pipeline for \textbf{SS2V} method. Yellow color highlights the initial human-annotated BB. Green color highlights the candidate BBs generated from the predicted masks. Blue color highlights synthetic BBs generated by the \textbf{SS2V} method.} 
\label{fig:ss2v}
\end{figure}

We next introduce the \textbf{SS2V} method for pixel-level segmentation of a 3D volume only using a single 2D BB annotation. 
The workflow of \textbf{SS2V} is depicted in Fig.\ \ref{fig:ss2v}. 

1. We first select a 2D slice \( K \)containing the target. A manual BB, labelled as \textbf{Box K}, is provided. While in theory our method can start from any slice containing the target, we begin with the central slice, as this typically offers a more representative view of the anatomy than a far-off side.

2. Using \textbf{Box K} as the initial BB, the segmentation \textbf{Pred K} is derived using our FNPC method. Based on \textbf{Pred K}, we generate a tightly fitting BB, \textbf{CBox K}, to be used as the candidate BB for the next slice. 

3. We refine the \textbf{CBox K} based on \textbf{Box K} to generate the BB (\textbf{Box K+1} or \textbf{Box K-1}) for the neighboring slices.

4. Steps 2 and 3 are iteratively applied to produce BBs for subsequent slices until the whole volume is segmented.

We assume that transitions between neighboring slices should be smooth. We enforce this by comparing \textbf{CBox K} and \textbf{Box K} and restricting the movement of each corner by a threshold \( T_{B} \) in Step 3. For example, for the lower-left corner of \textbf{Box K+1}, we use \( xmin_{B K+1} = xmin_{CB K} \) if \( |xmin_{CB K} - xmin_{B K}| \leq T_{B} \), and \( xmin_{B K+1} = xmin_{B K} \) otherwise.

\subsection{Datasets, preprocessing, and implementation details}
\label{subsec:data}
\textbf{\underline{Kidney:}} We use a dataset of free-hand kidney ultrasound images. It comprises 124 samples from 9 subjects, each 128x128 in dimension, with manual segmentations. The images are normalized to [0, 255] range. We compute \textbf{Fine BB} as the tightest BB of the pixel-level masks and randomly expanding the edges outwards by 0 to 2 pixels. \textbf{Medium BB} and \textbf{Coarse BB} are produced by similar expansions, ranging from 2 to 4 pixels and 4 to 6 pixels, respectively. 

\noindent
\textbf{\underline{Placenta:}} We use a  3D placenta ultrasound dataset from 4 subjects, with manual annotations.  The images are resized to 128x128x128 and normalized to [0, 255] range. We extract all 2D coronal slices containing placenta, for a total of 222 2D images. For brevity, we only show the \textbf{Fine BB} setting. For \textbf{SS2V}, we use the Fine BB of the central slice as initial BB.

\noindent
\textbf{\underline{Hyperparameter selection:}} We use the pretrained ViT-L SAM model\cite{kirillov2023segment} to obtain the initial SAM segmentation.  For kidneys, we randomly pick 14 images from one subject for hyperparameter setting. $T_{UC}$ is 0.9 for the Fine BB, 0.1 for the Medium BB and Coarse BB. For all stages, $M$ is 8, $N$ is 30, $T_{FNl}$ and $T_{FPl}$ are both 0, $T_{FNh}$ and $T_{FPh}$ are both 20. For the placenta segmentation and SS2V task, we extract the central slice of each subject for hyperparameter tuning. For both tasks, $T_{UC}$ is 0.2, $M$ is 4, $N$ is 30, $T_{FNl}$ and $T_{FPl}$ are both 70, $T_{FNh}$ and $T_{FPh}$ are both 200. For SS2V, the $T_{B}$ is 2.


\section{Results and Conclusion}

\begin{figure}[h]
    \centering
    \begin{minipage}[c]{0.4\linewidth}
    \centering
    \includegraphics[height=9cm]{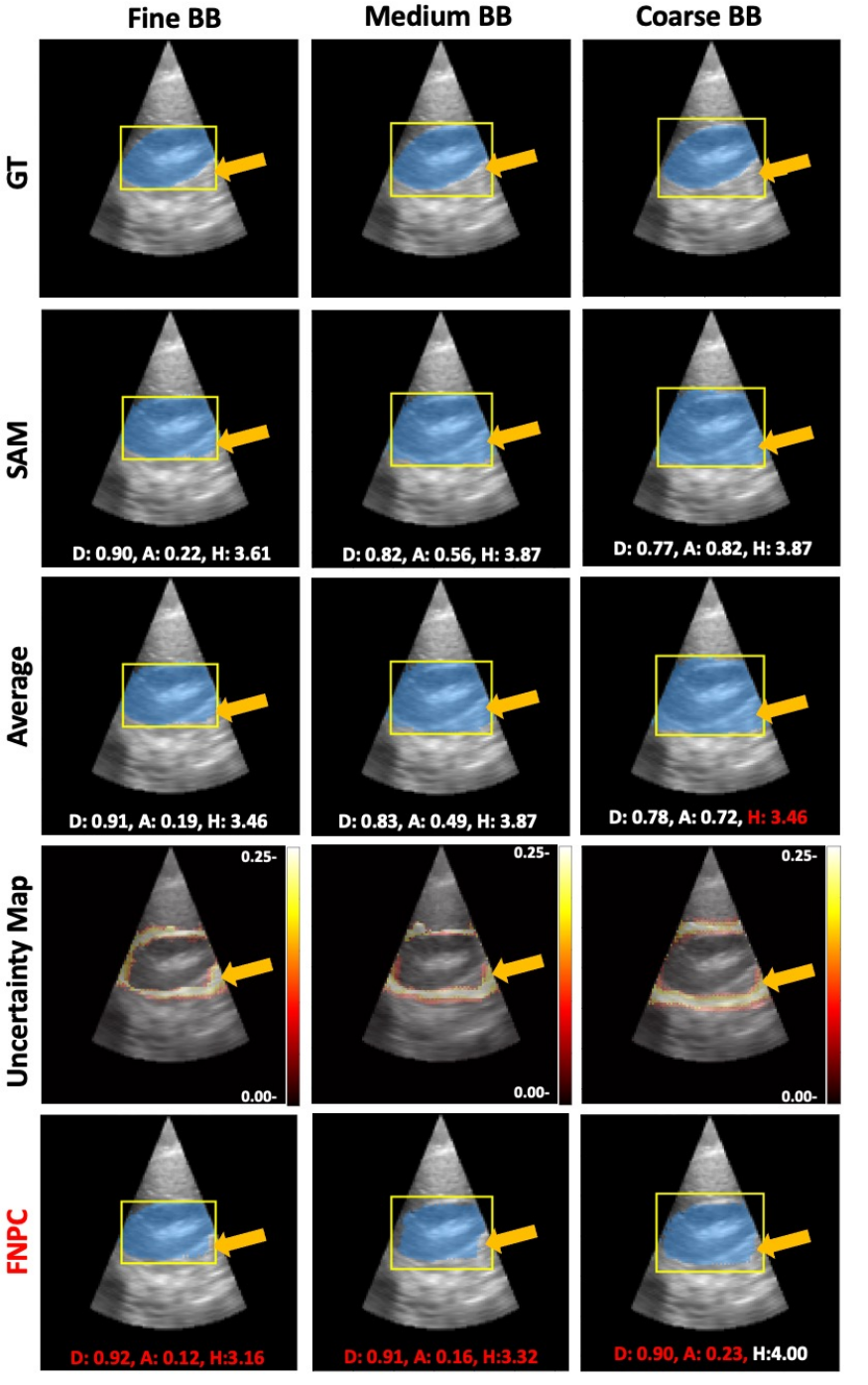}
    \end{minipage}
     \begin{minipage}[c]{0.55\linewidth}
     \centering
     \includegraphics[height=9cm]{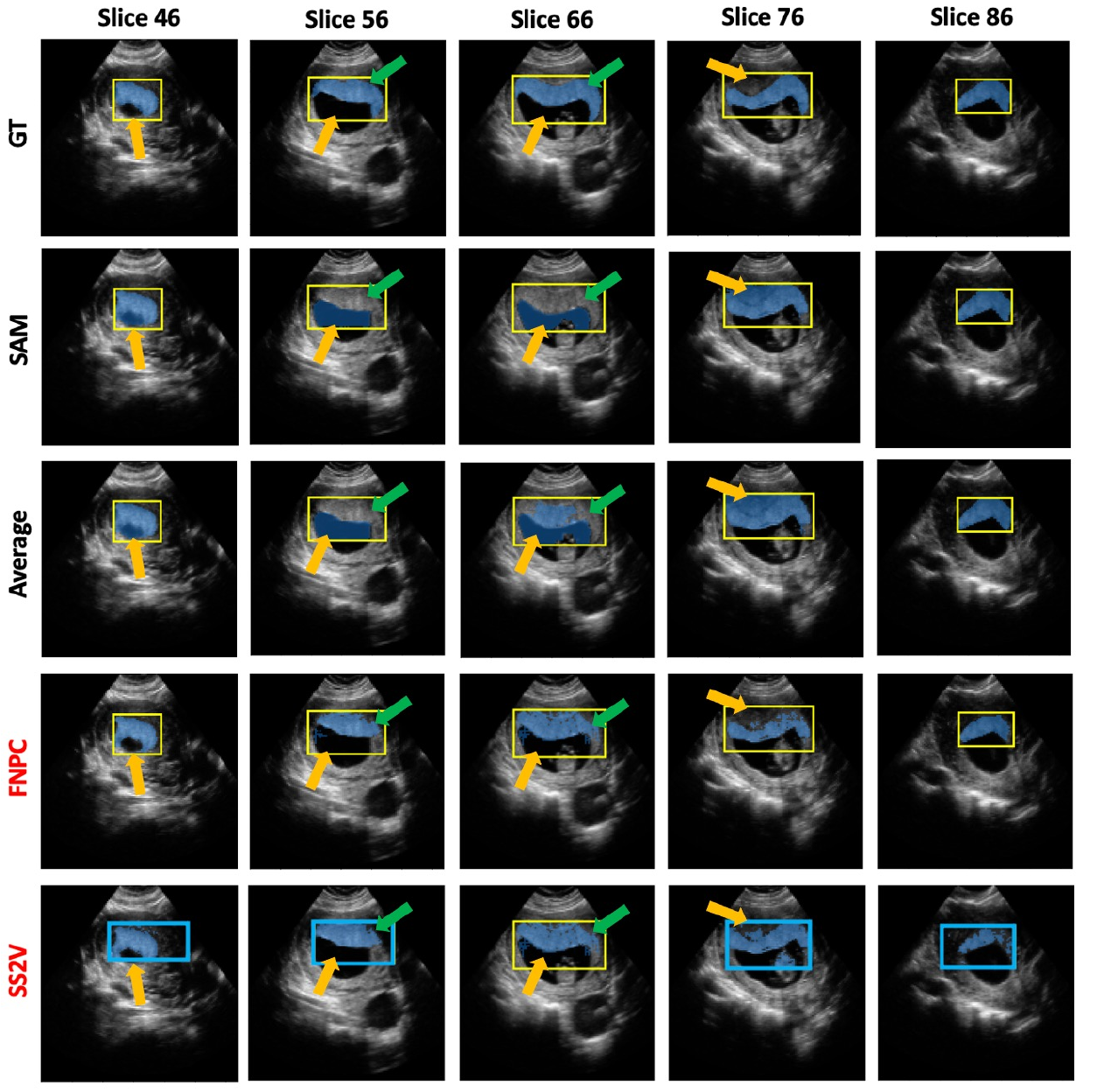}
     \end{minipage}
    \caption{Qualitative results. \textbf{Left,} kidney segmentation for three levels of prompt coarseness. D: Dice, A: ASSD, H: HD. Red font indicates the best performance.  \textbf{Right,} placenta segmentation with SS2V. Yellow: manual BBs, blue: synthesized BBs generated by SS2V. For \textbf{Left} and \textbf{Right}, green and orange arrows highlight the FNs and FPs improved by our proposed methods (highlighted in red). }
    \label{fig:qualitative}
\end{figure}

\textbf{\underline{Kidney:}} Left panels of Fig.\ \ref{fig:qualitative} and Table \ref{tab:quantitative} compare SAM, Average, and FNPC on kidney images under three levels of prompt coarseness, qualitatively and quantitatively. FNPC effectively eliminates the FP portions within the average predictions and SAM, and improves the Dice, ASSD, and HD results, across all BB coarseness levels. We observe that unlike SAM and average predictions, FNPC only shows a small deterioration in Dice and ASSD between fine and coarse prompts. This highlights FNPC's robustness to prompt coarseness. 

\begin{table}[h]
    \centering
    \begin{minipage}[c]{0.48\textwidth}
        \centering
        \scriptsize
        \begin{adjustbox}{minipage=\textwidth, valign=c, center}
        \begin{tabular}{|M{0.09\textwidth}|M{0.025\textwidth}|M{0.18\textwidth}|M{0.18\textwidth}|M{0.18\textwidth}|}
            \hline
            \textbf{Kidney} & BB & Dice$\uparrow$ & ASSD$\downarrow$ & HD$\downarrow$ \\
            \hline
             & F & 0.90$\pm$0.05 & 0.25$\pm$0.21 & 3.11$\pm$0.83 \\
            SAM & M & 0.87$\pm$0.06 & 0.35$\pm$0.29 & 3.26$\pm$0.81 \\
             & C & 0.77$\pm$0.57 & 0.83$\pm$0.37 & 3.67$\pm$0.67 \\
            \hline 
             & F & 0.90$\pm$0.05 & 0.24$\pm$0.20 & 3.12$\pm$0.83 \\
            Ave & M & 0.87$\pm$0.06 & 0.34$\pm$0.29 & 3.25$\pm$0.83 \\
             & C & 0.70$\pm$0.06 & 0.79$\pm$0.35 & 3.63$\pm$0.66 \\
            \hline
             & F & \textcolor{red}{0.90$\pm$0.04} & \textcolor{red}{0.20$\pm$0.17} & \textcolor{red}{3.09$\pm$0.82} \\
            FNPC & M  & \textcolor{red}{0.88$\pm$0.05} & \textcolor{red}{0.24$\pm$0.22} & \textcolor{red}{3.14$\pm$0.78} \\
             & C & \textcolor{red}{0.88$\pm$0.04} & \textcolor{red}{0.31$\pm$0.18} & \textcolor{red}{3.35$\pm$0.81} \\
            \hline
        \end{tabular}
        \end{adjustbox}
    \end{minipage}
    \begin{minipage}{0.48\textwidth}
        \centering
        \scriptsize
        \begin{tabular}{|M{0.12\textwidth}|M{0.025\textwidth}|M{0.2\textwidth}|M{0.2\textwidth}|M{0.2\textwidth}|}
            \hline
            \textbf{Placenta}& BB&Dice$\uparrow$ & ASSD$\downarrow$ & HF$\downarrow$ \\
            \hline
            SAM & F & $0.71 \pm 0.24$ & $1.15 \pm 2.02$ & $3.61 \pm 1.04$ \\
            Ave & F& $0.70 \pm 0.17$ & $1.21 \pm 0.58$ & $3.87 \pm 1.01$ \\
            FNPC & F& \xxx{$0.78 \pm 0.08$} & \xxx{$0.55 \pm 0.30$} & \xxx{$3.49 \pm 1.06$} \\
            SS2V & F& $0.72 \pm 0.17$ & $1.14 \pm 1.42$ & $3.90 \pm 1.06$ \\
            \hline
        \end{tabular}
    \end{minipage}
    \caption{Quantitative results. F, M, C represent Fine BB, Medium BB, and Coarse BB, respectively. \textbf{Left,} kidney. Red numbers indicate the best performance for each BB scenario. \textbf{Right,} placenta with SS2V. Red numbers indicate the best performance for each metric. }
    \label{tab:quantitative}
\end{table}

\noindent
\textbf{\underline{Placenta and SS2V:}} The right panel of Fig.\ \ref{fig:qualitative} presents the  results for 3D placenta segmentation using the SS2V method.  The top four rows depict 2D segmentation results, using a manual Fine BB annotation (yellow box). The FNPC method delivers overall superior segmentation results, with fewer FPs and FNs than the SAM and Average methods. Remarkably, SS2V showcases performance on par with FNPC across all slices, even though it only uses a single 2D Fine BB annotation (yellow box) for the entire 3D segmentation task. We observe that synthetic BBs (blue boxes) produced by SS2V are less precise than the manual Fine Box annotations, especially as we move further away from the reference slice 66. Nonetheless, even with these coarse annotations, SS2V manages to yield segmentations that rival those produced by methods using a new manual BB annotations for each slice. This performance leverages FNPC's robustness to BB variations. The right panel of Table\ \ref{tab:quantitative} provides quantitative results, illustrating that FNPC has better performance than SAM and Ave. SS2V's outcomes are closely aligned with those of FNPC, showcasing excellent extension to 3D segmentation tasks.

\noindent
\textbf{\underline{Conclusion:}}
We introduced a test-phase prompt augmentation method to adapt SAM to challenging medical image segmentation tasks, specifically for ultrasound images marked by low contrast and noise. This method, leveraging multi-box prompt augmentation and aleatoric uncertainty thresholding, aims to mitigate SAM's FN and FP predictions without requiring time-consuming pixel-level annotations. Our evaluation on two ultrasound datasets showcases substantial improvements in SAM's performance and robustness to prompt coarseness. We further proposed SS2V to produce 3D segmentations from a single 2D BB input, with excellent results.
We recognize, however, that continued exploration and optimization are required for dealing with increasingly complex and variable data.

\acknowledgments
This work is supported, in part, by NIH R01HD109739 and NIH R01HL156034. Here we also express our thanks for the computational resource support by Advanced Computing Center for Research and Education (ACCRE) at Vanderbilt University.  
%


\bibliographystyle{spiebib} 

\end{document}